\documentclass[conference]{IEEEtran}
\IEEEoverridecommandlockouts
\usepackage{graphicx}
\usepackage{setspace}
\usepackage{amsmath}
\usepackage{dsfont}
\usepackage{amsfonts}
\usepackage{lipsum}
\usepackage{multicol}
\usepackage{float}
\usepackage{bm}
\usepackage{color}
\usepackage[dvipsnames]{xcolor}
\usepackage{etoolbox}
\usepackage{soul}
\usepackage[linesnumbered,ruled,vlined]{algorithm2e}
\usepackage{mathtools}
\usepackage{caption}
\usepackage{amsthm}
\usepackage{adjustbox}
\usepackage{afterpage}
\usepackage{multirow}
\usepackage{mathrsfs}
\usepackage{hyperref}
\usepackage{oplotsymbl}
\usepackage{booktabs}
\usepackage{array}
\usepackage{xcolor}
\usepackage{amssymb}
\usepackage{pifont}
\usepackage[table]{xcolor}
\usepackage[caption = false,subrefformat=parens,labelformat=parens]{subfig}

\definecolor{myteal}{HTML}{007C97}
\definecolor{mygreen}{HTML}{009A66}
\definecolor{myred}{HTML}{ED1C24}

\title{Beyond Fixed Rounds: Data-Free Early Stopping for Practical Federated Learning

\author{
Youngjoon Lee\textsuperscript{\rm 1}, Hyukjoon Lee\textsuperscript{\rm 2}, Seungrok Jung\textsuperscript{\rm 2}, Andy Luo\textsuperscript{\rm 2}, Jinu Gong\textsuperscript{\rm 3}, Yang Cao\textsuperscript{\rm 4}, and Joonhyuk Kang\textsuperscript{\rm 1} \\
\textsuperscript{\rm 1}School of Electrical Engineering, KAIST, South Korea\\
\textsuperscript{\rm 2}AI Group, AMD, United States\\
\textsuperscript{\rm 3}Department of Applied AI, Hansung University, South Korea\\
\textsuperscript{\rm 4}Department of Computer Science, Institute of Science Tokyo, Japan\\
Email: yjlee22@kaist.ac.kr, jkang@kaist.ac.kr
}
}

\begin{document}

\maketitle

\begin{abstract}
Federated Learning (FL) facilitates decentralized collaborative learning without transmitting raw data. 
However, reliance on fixed global rounds or validation data for hyperparameter tuning hinders practical deployment by incurring high computational costs and privacy risks. 
To address this, we propose a data-free early stopping framework that determines the optimal stopping point by monitoring the task vector's growth rate using only server-side parameters. 
The numerical results on skin lesion/blood cell/colon pathology classification demonstrate that 
our approach is comparable to the validation-based early stopping across various state-of-the-art FL methods.
In particular, the proposed framework requires an average of 45/12/31 (skin lesion/blood cell/colon pathology) additional rounds to achieve over 12.3\%/8.9\%/3.9\% higher performance than early stopping based on validation data.
Moreover, the proposed framework requires only 9/8/14 additional rounds to screen bad configurations, which is less than 3\% of the fixed-round budget.
To the best of our knowledge, this is the first work to propose a data-free early stopping framework for FL methods.
Our code is available at \href{https://github.com/yjlee22/earlystop}{this open repository}.
\end{abstract}

\section{Introduction}
\label{sec:intro}
Deep learning has driven significant advancements in medical imaging, utilizing large-scale datasets to achieve remarkable diagnostic performance \cite{wang2024comprehensive,bisio2025ai}. 
However, the deployment of AI is strictly limited by stringent privacy regulations that prohibit the centralization of sensitive patient data \cite{peloquin2020disruptive}.
To overcome this barrier, Federated Learning (FL) \cite{mcmahan2017communication} has emerged as a promising decentralized paradigm that facilitates collaborative learning without transmitting raw data \cite{abouelmehdi2018big}. 
By ensuring that data remain localized at their source, FL preserves data sovereignty and strictly adheres to institutional governance and ethical standards \cite{li2020federated}. 
Moreover, FL enables robust and generalizable learning in medical AI by effectively leveraging cross-institutional data diversity \cite{antunes2022federated}.
Thus, FL provides a scalable and privacy-preserving solution well suited for the secure development of collaborative medical AI systems \cite{joshi2022federated}.

Recent FL methods have evolved to enhance convergence stability and performance by refining optimization techniques at the client or server level \cite{bakas2026federated}.
The foundational approaches are typically grounded in Stochastic Gradient Descent (SGD) \cite{simeone2022machine}, where FL methods like FedAvg, FedProx \cite{li2018federated}, SCAFFOLD \cite{karimireddy2020scaffold}, and FedDyn \cite{acarfederated} regulate updates to stabilize the learning process.
Subsequently, advanced FL methods have shifted towards Sharpness-Aware Minimization (SAM) \cite{foretsharpness}, which seeks flat minima rather than merely minimizing loss values.
This paradigm has led to the development of SAM-based FL methods—including FedSAM \cite{qu2022generalized}, FedSpeed \cite{sunfedspeed}, FedSMOO \cite{sun2023dynamic}, FedGamma \cite{10269141}, FedLESAM \cite{FedLESAM}, and FedWMSAM \cite{FedWMSAM}—that smooth the optimization trajectory.

\begin{figure}[t]
     \centering
     \includegraphics[width=\columnwidth]{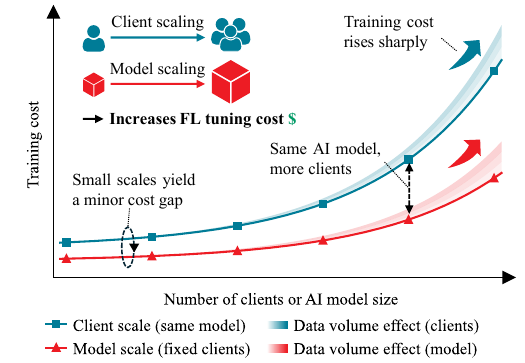}
     \caption{
    Illustration of the training cost challenge in FL hyperparameter tuning.
    Different scaling factors, including the number of clients and AI model size, can substantially increase the tuning cost.
    Note that client scaling can lead to sharper cost growth by increasing the server--client communication cost.
    }
     \label{fig:intro}
\end{figure}

\begin{figure*}[t]
\centering
\includegraphics[scale=1.33]{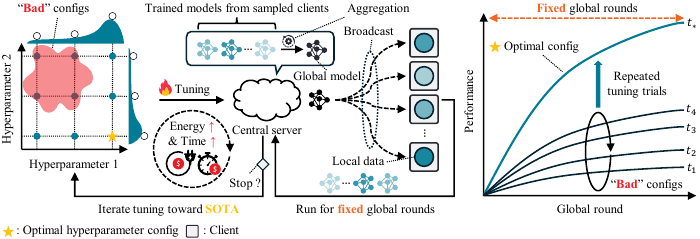}
\caption{Illustration of resource inefficiency in FL hyperparameter tuning. Since standard FL protocols use a fixed number of global rounds, \textcolor{myred}{\textbf{bad}} configurations waste computational and communication resources.
This motivates the need for early stopping in FL for scalable and practical deployment.
}
\label{fig:problem}
\end{figure*}

Despite the strong performance of recent FL methods, a key limitation remains in their reliance on a fixed number of global rounds for training \cite{lee2025revisit}.
This limitation becomes more critical in practical deployment, where FL requires exploring a large hyperparameter space over different FL methods, data distributions, and training configurations \cite{lee2025debunking}.
Moreover, the tuning cost increases sharply with scaling factors such as the number of clients and the AI model size \cite{10038639}, as illustrated in Fig.~\ref{fig:intro}.
In particular, client scaling amplifies the communication overhead, leading to steeper cost growth \cite{khodak2021federated}.
Although many configurations yield similar objective values, they can lead to different outcomes under fixed-round training.
This issue is especially severe for bad configurations under fixed-round training, where computational and communication waste becomes more pronounced, as illustrated in Fig.~\ref{fig:problem}.

In this work, we propose a novel data-free early stopping framework that determines when to stop training using only the global model parameters at the server.
Note that, unlike existing approaches relying on validation signals \cite{10643330,11400880}, our framework adopts a purely model-driven stopping criterion.
By avoiding the need for validation data, our framework strictly adheres to the FL paradigm of model-only transmission \cite{kairouz2021advances,rieke2020future,rauniyar2023federated}. 
We show that our approach seamlessly integrates with 10 state-of-the-art FL methods and remains robust across medical imaging datasets. 
Moreover, our framework maintains consistent stability under various non-IID distributions, effectively handling data heterogeneity. 
The experiments validate that our framework achieves generalization performance comparable to that of approaches relying on validation data \cite{prechelt2002early,yao2007early,xu2018splitting}.
Main contributions of this paper are as follows:

\begin{itemize}
    \item We propose a data-free early stopping framework for FL that identifies the stopping point using only server-side global model parameters.
    \item We introduce a task-vector-based criterion that captures training stability by monitoring the growth rate of accumulated parameter displacement.
    \item We demonstrate that the proposed framework achieves validation-level performance on medical classification tasks across diverse FL methods.
    \item In addition, we show that the proposed framework can effectively screen out bad configurations early, reducing the computational waste of fixed-round training.
\end{itemize}

The remainder of this paper is organized as follows.  
In Section~\ref{sec:main}, we formulate the problem and describe the proposed data-free early stopping framework based on task-vector dynamics.  
In Section~\ref{sec:experiment}, we present numerical results and analyze the effectiveness of the proposed framework under diverse FL settings.  
Finally, Section~\ref{sec:conclusion} concludes with remarks.

\begin{figure*}[t]
\centering
\includegraphics[scale=1.33]{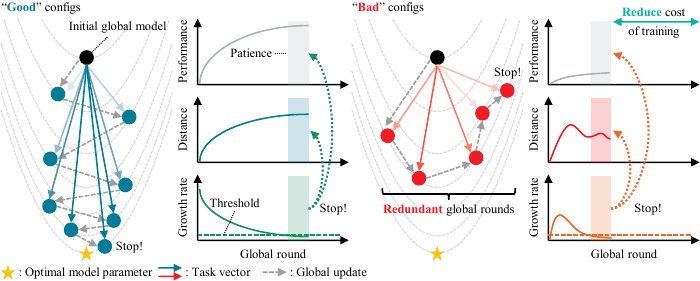}
\caption{Illustration of the proposed data-free early stopping framework.
The server monitors the growth rate of the task vector using only global model parameters and stops training once the growth rate falls below the threshold. Here, the color intensity of the task vector reflects its increasing magnitude, and vice versa.
}
\label{fig:method}
\end{figure*}

\section{Problem and Model}\label{sec:main}
\subsection{Federated Setting}
We consider a federated network comprising a central server and $N$ clients, designed for medical image classification tasks.
The system aims to optimize the global model parameters $\theta$ by minimizing the global objective function as:
\begin{equation}
F(\theta) \triangleq \frac{1}{N} \sum_{n=1}^{N} F_n(\theta),
\end{equation}
where $F_n(\theta)$ denotes the local objective function calculated over the private local dataset $\mathcal{D}_n$ of the $n$-th client.
Note that the global dataset is defined as $\mathcal{D}=\bigcup_{n=1}^{N}\mathcal{D}_n$.
To simulate the non-IID nature, we consider three non-IID partitioning types following \cite{li2022federated}: \emph{label skew} (Dirichlet, Pathological) and \emph{quantity skew}.
For all considered distributions, the specific data allocation across clients is determined by a coefficient $c$.

For \emph{label skew} (Dirichlet), the client-wise allocation proportions for samples of class $z$ are sampled as:
\begin{equation}
    \boldsymbol{p}^{(z)} \sim \operatorname{Dir}_{N}(c), \quad
    |\mathcal{D}_n^{(z)}| \approx p_n^{(z)}|\mathcal{D}^{(z)}|,
\end{equation}
where $\mathcal{D}^{(z)} \subseteq \mathcal{D}$ and $\mathcal{D}_n^{(z)} \subseteq \mathcal{D}_n$ denote the samples of class $z$ in the global and local datasets, respectively.
Note that a smaller $c$ induces stronger label imbalance across clients.
By contrast, \emph{label skew} (Pathological) restricts each local dataset to samples from only $c$ distinct classes:
\begin{equation}
    \bigl|\{z:\mathcal{D}_n^{(z)} \neq \emptyset\}\bigr|=c, \quad
    \forall n \in \{1,\ldots,N\}.
\end{equation}
Thus, a smaller $c$ limits each client to fewer classes, resulting in stronger label imbalance.

Meanwhile, \emph{quantity skew} varies the local dataset sizes through client-wise proportions sampled as:
\begin{equation}
    \boldsymbol{q} \sim \operatorname{Dir}_{N}(c), \quad
    |\mathcal{D}_n| \approx q_n |\mathcal{D}|.
\end{equation}
As with \emph{label skew} (Dirichlet), a smaller $c$ induces stronger quantity imbalance across clients.

\subsection{Proposed Framework: Data-Free Early Stopping}
\label{sec:proposed}
We propose a novel early stopping framework that utilizes task vector characteristics \cite{li2025task,zhou2025task} using only server-side global parameters, without relying on validation data.
In detail, the FL process starts at global round $r=1$ with an initialized global model $\boldsymbol{\theta}_0$.
At each round $r \geq 1$, randomly sampled $M$ clients compute local updates through client-side optimization \textsc{ClientOpt}($\cdot$). 
Then, the central server aggregates these local updates through server-side optimization \textsc{ServerOpt}($\cdot$) to obtain the global model $\boldsymbol{\theta}_r$.
Note that $\Phi_{\mathrm{client}}$ and $\Phi_{\mathrm{server}}$ denote the key hyperparameters of \textsc{ClientOpt}($\cdot$) and \textsc{ServerOpt}($\cdot$), respectively, for each FL method \cite{lee2023fast}.
We define the \emph{global task vector} $\mathbf{v}_r \in \mathbb{R}^d$ as the cumulative displacement from the initialization:
\begin{equation}\label{eq1}
\mathbf{v}_r := \boldsymbol{\theta}_r - \boldsymbol{\theta}_0
= \sum_{k=1}^{r} \left( \boldsymbol{\theta}_{k} - \boldsymbol{\theta}_{k-1} \right).
\end{equation}
As training progresses, the global model moves away from the initialization, inducing an increasing task-specific displacement in the parameter space \cite{ilharcoediting}.
From an optimization perspective, each global update can be interpreted as a fine-tuning step resulting from the coupled dynamics of \textsc{ClientOpt}($\cdot$) and \textsc{ServerOpt}($\cdot$).
Under standard smoothness assumptions, $\mathbf{v}_r$ approximates the accumulated gradient flow as:
\begin{equation}\label{eq2}
\mathbf{v}_r \approx
- \sum_{k=1}^{r} \gamma_k \nabla F(\boldsymbol{\theta}_{k-1}).
\end{equation}
Here, $\gamma_k$ represents the effective step size determined by the local learning rate, the number of local steps, and the aggregation scaling.
Since FL satisfies stationarity conditions, i.e.,
$\lim_{r \to \infty} \| \nabla F(\boldsymbol{\theta}_r) \|_2 = 0$,
the growth of $\mathbf{v}_r$ diminishes as training stabilizes.
Thus, the accumulated optimization distance,
$\delta_r := \| \mathbf{v}_r \|_2$,
gradually converges to a stable value.

To capture this training stabilization behavior, we introduce the growth rate $g_r$, which quantifies the magnitude of the relative change in the accumulated distance:
\begin{equation}\label{eq3}
    g_r = \frac{|\delta_r - \delta_{r-1}|}{\delta_{r-1}}, \quad r \geq 2.
\end{equation}
As the learning trajectory stabilizes, $g_r$ tends to decrease, indicating that later global updates contribute marginally to the overall displacement.
This behavior reflects the onset of saturation of the parameter-space trajectory.

\begin{figure}[t]
     \centering
     \includegraphics[width=\columnwidth]{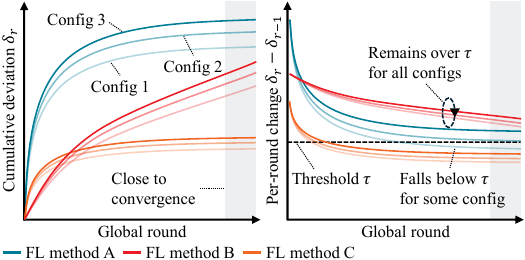}
     \caption{
    Illustration of why the per-round change $\delta_r - \delta_{r-1}$ is unsuitable for stopping across FL methods and configurations.
    }
    \label{fig:reason}
\end{figure}

We adopt $g_r$ rather than the simple difference $\delta_r - \delta_{r-1}$ for the following reasons.
In detail, by the triangle inequality,
\begin{equation}
\delta_r \;\lesssim\; \sum_{k=1}^{r} \gamma_k \, \| \nabla F(\boldsymbol{\theta}_{k-1}) \|_2,
\end{equation}
so the scale of $\delta_r$ is governed by $\gamma_k$, and no single $\tau$ works across the hyperparameter settings of different FL methods.
By the same reasoning, the per-round change likewise satisfies
\begin{equation}
\delta_r - \delta_{r-1} \;\lesssim\; \gamma_r \, \| \nabla F(\boldsymbol{\theta}_{r-1}) \|_2,
\end{equation}
so the criterion $\delta_r - \delta_{r-1} \leq \tau$ may fail for any fixed $\tau > 0$ even when the global model is close to convergence.
Thus, the scale-dependent difference $\delta_r-\delta_{r-1}$ is unsuitable as a stopping indicator across FL methods, as shown in Fig.~\ref{fig:reason}.

To align with the validation-based early stopping, the proposed criterion is restricted to two hyperparameters: a sensitivity threshold $\tau$ and a patience parameter $\rho$.
In particular, we define a recursive saturation counter $\kappa_r$ as:
\begin{equation}\label{eq4}
    \kappa_r = \mathbb{I}(g_r < \tau) \cdot (\kappa_{r-1} + 1), \quad \kappa_1 = 0,
\end{equation}
where $\mathbb{I}(\cdot)$ denotes the indicator function.
The federated training process is stopped at the round $r^*$ satisfying:
\begin{equation}\label{eq5}
    r^* = \min \{ r \geq 2 \mid \kappa_r \geq \rho \}.
\end{equation}
The overall proposed framework and procedure are shown in Fig.~\ref{fig:method} and Algorithm~\ref{alg}, respectively.

\begin{algorithm}[t]
\caption{Proposed Framework}
\label{alg}
\SetKwProg{Fn}{Function}{:}{}
\SetKwFunction{FCheckEarlyStop}{CheckEarlyStop}
\SetKwFunction{FClientOpt}{ClientOpt}
\SetKwFunction{FServerOpt}{ServerOpt}
\SetKwInOut{Input}{Input}
\SetKwInOut{Output}{Output}
\Input{Initial model $\boldsymbol{\theta}_0$, threshold $\tau$, patience $\rho$}
\Output{Stopping-round global model $\boldsymbol{\theta}_{r^\star}$}
\BlankLine
\For{$r=1$ \KwTo $R$}{
    \ForEach{selected clients \textbf{in parallel}}{
        $\boldsymbol{\theta}_r^{m} \leftarrow$ \textsc{ClientOpt}({$\boldsymbol{\theta}_{r-1}$})\;
    }
    $\boldsymbol{\theta}_{r} \leftarrow$ \textsc{ServerOpt}({$\{\boldsymbol{\theta}_r^{m}\}_{m=1}^{M}$})\;
    $r^\star \leftarrow$ \textsc{CheckEarlyStop}({$\boldsymbol{\theta}_{r}, r$})\;
    \lIf{$r^\star > 0$}{\Return $\boldsymbol{\theta}_{r^\star}$}
}
\Return $\boldsymbol{\theta}_R$\;
\BlankLine
\Fn{\textsc{CheckEarlyStop}{($\boldsymbol{\theta}_r, r$)}}{
    \textcolor{myteal}{$\triangleright$ \textbf{Stopping criterion}}\;
    Compute $\mathbf{v}_r = \sum_{k=1}^{r} \left( \boldsymbol{\theta}_{k} - \boldsymbol{\theta}_{k-1} \right)$\\
    \If{$r \ge 2$}{
        Compute $g_r = \frac{|\delta_r - \delta_{r-1}|}{\delta_{r-1}}$\\
        Update $\kappa_r=\mathbb{I}(g_r < \tau) \cdot (\kappa_{r-1} + 1)$\\
        \lIf{$\kappa_r \ge \rho$}{\Return $r$}
    }
    \Return $0$\;
}
\end{algorithm}

\subsection{Analysis of Proposed Criterion}
We analyze the behavior of the proposed criterion under representative training dynamics.
Let $\boldsymbol{u}_r$ denote the global model update performed at round $r$:
\begin{equation}
\boldsymbol{u}_r := \boldsymbol{\theta}_r-\boldsymbol{\theta}_{r-1}.
\end{equation}
Since $\boldsymbol{u}_r=\mathbf{v}_r-\mathbf{v}_{r-1}$, the reverse triangle inequality gives
\begin{equation}
|\delta_r-\delta_{r-1}| = \left|\|\mathbf{v}_r\|_2-\|\mathbf{v}_{r-1}\|_2\right| \leq \|\boldsymbol{u}_r\|_2.
\end{equation}
Accordingly, applying the above inequality yields
\begin{equation}
g_r = \frac{|\delta_r-\delta_{r-1}|}{\delta_{r-1}} \leq \frac{\|\boldsymbol{u}_r\|_2}{\delta_{r-1}}.
\end{equation}
Thus, the criterion becomes small when the current update is negligible relative to the accumulated displacement.

\textbf{Good Configurations.}
For a convergent configuration \cite{10833754}, the global update can be viewed as an effective descent step:
\begin{equation}
\boldsymbol{u}_r \approx -\gamma_r \nabla F(\boldsymbol{\theta}_{r-1}).
\end{equation}
Accordingly, its magnitude is approximately given by
\begin{equation}
\|\boldsymbol{u}_r\|_2 \approx \gamma_r\|\nabla F(\boldsymbol{\theta}_{r-1})\|_2.
\end{equation}
Combining this relation with the upper bound on $g_r$ yields
\begin{equation}
g_r \leq \frac{\|\boldsymbol{u}_r\|_2}{\delta_{r-1}} \approx \frac{\gamma_r\|\nabla F(\boldsymbol{\theta}_{r-1})\|_2}{\delta_{r-1}}.
\end{equation}
As the trajectory approaches a stationary region with a bounded effective step size, we obtain
\begin{equation}
\|\nabla F(\boldsymbol{\theta}_{r-1})\|_2 \rightarrow 0 \quad \Longrightarrow \quad \|\boldsymbol{u}_r\|_2 \rightarrow 0 \quad \Longrightarrow \quad g_r \rightarrow 0.
\end{equation}
This implies that a small $g_r$ indicates that subsequent rounds add marginal displacement to the task-specific trajectory.

\textbf{Bad Configurations.}
In an ineffective configuration, the global update may remain bounded by a small update scale $\varepsilon>0$ over consecutive rounds:
\begin{equation}
\|\boldsymbol{u}_r\|_2 \leq \varepsilon, \qquad \varepsilon>0.
\end{equation}
Using this relation with the upper bound on $g_r$ gives
\begin{equation}
g_r \leq \frac{\|\boldsymbol{u}_r\|_2}{\delta_{r-1}} \leq \frac{\varepsilon}{\delta_{r-1}}.
\end{equation}
For a relatively large threshold $\tau>0$, suppose that the accumulated displacement satisfies
\begin{equation}
\delta_{r-1}>\frac{\varepsilon}{\tau}.
\end{equation}
Since $\delta_{r-1}>0$ and $\tau>0$, this condition can be rewritten as:
\begin{equation}
\frac{\varepsilon}{\delta_{r-1}}<\tau.
\end{equation}
Therefore, by the derived bounds, $g_r$ is bounded as follows:
\begin{equation}
g_r \leq \frac{\varepsilon}{\delta_{r-1}} < \tau.
\end{equation}
This allows a relatively large $\tau$ to stop configurations early when additional movement becomes negligible.

\begin{table*}[t]
\centering
\footnotesize
\renewcommand{\arraystretch}{1.6}
\setlength{\tabcolsep}{2pt}
\definecolor{arrowup}{HTML}{007C97}
\definecolor{arrowdown}{HTML}{ED1C24}
\caption{Top-1 test accuracy (\%) across datasets and \emph{data skew} settings, where $L_d$, $L_p$, and $Q$ denote \emph{label skew} (Dirichlet, $c=0.1$), \emph{label skew} (Pathological, $c=2.0$), and \emph{quantity skew} ($c=0.1$), respectively. Each cell shows the best result over $\log\Phi \in \{\text{-}1, \text{-}2, \text{-}3\}$ with the selected $\Phi$ value in parentheses. 
The best results per row are \underline{\textbf{bold and underlined}}. 
Note that {\textcolor{arrowup}{$\blacktriangle$}} indicates the \emph{data skew} setting under which each method achieves its highest accuracy within a dataset; {\textcolor{arrowdown}{$\blacktriangledown$}} otherwise.
}
\begin{tabular}{cccccccccccc}
\toprule
 & \emph{Skew} & FedAvg & FedProx & FedDyn & SCAFFOLD & FedSAM & FedSpeed & FedSMOO & FedGamma & FedLESAM & FedWMSAM \\
\midrule
\multirow{3}{*}{\rotatebox{90}{Skin lesion}} & $L_d$ & $66.12\,\textcolor{arrowdown}{\blacktriangledown}$ & $67.05\,\textcolor{arrowdown}{\blacktriangledown}\;(\text{-}2)$ & $\underline{\mathbf{69.60\,\textcolor{arrowup}{\blacktriangle}\;(\text{-}2)}}$ & $68.12\,\textcolor{arrowdown}{\blacktriangledown}$ & $68.57\,\textcolor{arrowdown}{\blacktriangledown}\;(\text{-}2)$ & $69.42\,\textcolor{arrowdown}{\blacktriangledown}\;(\text{-}2)$ & $69.53\,\textcolor{arrowup}{\blacktriangle}\;(\text{-}2)$ & $68.22\,\textcolor{arrowdown}{\blacktriangledown}\;(\text{-}3)$ & $66.63\,\textcolor{arrowdown}{\blacktriangledown}\;(\text{-}2)$ & $68.15\,\textcolor{arrowdown}{\blacktriangledown}\;(\text{-}3)$ \\
 & $L_p$ & $66.76\,\textcolor{arrowup}{\blacktriangle}$ & $68.47\,\textcolor{arrowup}{\blacktriangle}\;(\text{-}3)$ & $67.74\,\textcolor{arrowdown}{\blacktriangledown}\;(\text{-}2)$ & $68.36\,\textcolor{arrowup}{\blacktriangle}$ & $\underline{\mathbf{70.76\,\textcolor{arrowup}{\blacktriangle}\;(\text{-}2)}}$ & $69.46\,\textcolor{arrowup}{\blacktriangle}\;(\text{-}2)$ & $68.88\,\textcolor{arrowdown}{\blacktriangledown}\;(\text{-}2)$ & $69.34\,\textcolor{arrowup}{\blacktriangle}\;(\text{-}3)$ & $68.03\,\textcolor{arrowup}{\blacktriangle}\;(\text{-}2)$ & $68.18\,\textcolor{arrowup}{\blacktriangle}\;(\text{-}3)$ \\
 & $Q$ & $55.72\,\textcolor{arrowdown}{\blacktriangledown}$ & $57.32\,\textcolor{arrowdown}{\blacktriangledown}\;(\text{-}2)$ & $\underline{\mathbf{64.16\,\textcolor{arrowdown}{\blacktriangledown}\;(\text{-}2)}}$ & $25.77\,\textcolor{arrowdown}{\blacktriangledown}$ & $57.26\,\textcolor{arrowdown}{\blacktriangledown}\;(\text{-}2)$ & $61.24\,\textcolor{arrowdown}{\blacktriangledown}\;(\text{-}2)$ & $62.53\,\textcolor{arrowdown}{\blacktriangledown}\;(\text{-}2)$ & $27.06\,\textcolor{arrowdown}{\blacktriangledown}\;(\text{-}3)$ & $55.85\,\textcolor{arrowdown}{\blacktriangledown}\;(\text{-}1)$ & $19.65\,\textcolor{arrowdown}{\blacktriangledown}\;(\text{-}1)$ \\
\midrule
\multirow{3}{*}{\rotatebox{90}{Blood cell}} & $L_d$ & $97.44\,\textcolor{arrowdown}{\blacktriangledown}$ & $97.69\,\textcolor{arrowdown}{\blacktriangledown}\;(\text{-}3)$ & $97.81\,\textcolor{arrowdown}{\blacktriangledown}\;(\text{-}3)$ & $97.33\,\textcolor{arrowdown}{\blacktriangledown}$ & $\underline{\mathbf{98.24\,\textcolor{arrowdown}{\blacktriangledown}\;(\text{-}2)}}$ & $97.84\,\textcolor{arrowdown}{\blacktriangledown}\;(\text{-}2)$ & $97.87\,\textcolor{arrowdown}{\blacktriangledown}\;(\text{-}2)$ & $98.00\,\textcolor{arrowdown}{\blacktriangledown}\;(\text{-}2)$ & $97.69\,\textcolor{arrowdown}{\blacktriangledown}\;(\text{-}3)$ & $98.14\,\textcolor{arrowup}{\blacktriangle}\;(\text{-}3)$ \\
 & $L_p$ & $97.61\,\textcolor{arrowdown}{\blacktriangledown}$ & $97.79\,\textcolor{arrowdown}{\blacktriangledown}\;(\text{-}2)$ & $97.87\,\textcolor{arrowdown}{\blacktriangledown}\;(\text{-}2)$ & $97.67\,\textcolor{arrowup}{\blacktriangle}$ & $\underline{\mathbf{98.15\,\textcolor{arrowdown}{\blacktriangledown}\;(\text{-}2)}}$ & $98.04\,\textcolor{arrowdown}{\blacktriangledown}\;(\text{-}2)$ & $97.98\,\textcolor{arrowdown}{\blacktriangledown}\;(\text{-}2)$ & $98.05\,\textcolor{arrowup}{\blacktriangle}\;(\text{-}2)$ & $97.66\,\textcolor{arrowdown}{\blacktriangledown}\;(\text{-}1)$ & $97.95\,\textcolor{arrowdown}{\blacktriangledown}\;(\text{-}3)$ \\
 & $Q$ & $98.02\,\textcolor{arrowup}{\blacktriangle}$ & $98.02\,\textcolor{arrowup}{\blacktriangle}\;(\text{-}2)$ & $98.16\,\textcolor{arrowup}{\blacktriangle}\;(\text{-}3)$ & $25.55\,\textcolor{arrowdown}{\blacktriangledown}$ & $\underline{\mathbf{98.25\,\textcolor{arrowup}{\blacktriangle}\;(\text{-}2)}}$ & $98.17\,\textcolor{arrowup}{\blacktriangle}\;(\text{-}2)$ & $98.13\,\textcolor{arrowup}{\blacktriangle}\;(\text{-}2)$ & $31.59\,\textcolor{arrowdown}{\blacktriangledown}\;(\text{-}3)$ & $98.07\,\textcolor{arrowup}{\blacktriangle}\;(\text{-}1)$ & $26.59\,\textcolor{arrowdown}{\blacktriangledown}\;(\text{-}1)$ \\
\midrule
\multirow{3}{*}{\rotatebox{90}{Colon path.}} & $L_d$ & $92.86\,\textcolor{arrowdown}{\blacktriangledown}$ & $93.52\,\textcolor{arrowdown}{\blacktriangledown}\;(\text{-}2)$ & $93.66\,\textcolor{arrowdown}{\blacktriangledown}\;(\text{-}3)$ & $90.88\,\textcolor{arrowup}{\blacktriangle}$ & $\underline{\mathbf{94.31\,\textcolor{arrowdown}{\blacktriangledown}\;(\text{-}1)}}$ & $93.55\,\textcolor{arrowdown}{\blacktriangledown}\;(\text{-}3)$ & $93.59\,\textcolor{arrowup}{\blacktriangle}\;(\text{-}2)$ & $93.04\,\textcolor{arrowup}{\blacktriangle}\;(\text{-}2)$ & $93.47\,\textcolor{arrowup}{\blacktriangle}\;(\text{-}2)$ & $78.59\,\textcolor{arrowup}{\blacktriangle}\;(\text{-}3)$ \\
 & $L_p$ & $93.76\,\textcolor{arrowup}{\blacktriangle}$ & $93.37\,\textcolor{arrowdown}{\blacktriangledown}\;(\text{-}2)$ & $92.26\,\textcolor{arrowdown}{\blacktriangledown}\;(\text{-}3)$ & $84.78\,\textcolor{arrowdown}{\blacktriangledown}$ & $\underline{\mathbf{94.16\,\textcolor{arrowdown}{\blacktriangledown}\;(\text{-}2)}}$ & $92.00\,\textcolor{arrowdown}{\blacktriangledown}\;(\text{-}3)$ & $92.89\,\textcolor{arrowdown}{\blacktriangledown}\;(\text{-}3)$ & $91.98\,\textcolor{arrowdown}{\blacktriangledown}\;(\text{-}2)$ & $93.30\,\textcolor{arrowdown}{\blacktriangledown}\;(\text{-}2)$ & $67.86\,\textcolor{arrowdown}{\blacktriangledown}\;(\text{-}3)$ \\
 & $Q$ & $93.49\,\textcolor{arrowdown}{\blacktriangledown}$ & $93.61\,\textcolor{arrowup}{\blacktriangle}\;(\text{-}2)$ & $93.78\,\textcolor{arrowup}{\blacktriangle}\;(\text{-}2)$ & $63.39\,\textcolor{arrowdown}{\blacktriangledown}$ & $\underline{\mathbf{95.68\,\textcolor{arrowup}{\blacktriangle}\;(\text{-}1)}}$ & $93.76\,\textcolor{arrowup}{\blacktriangle}\;(\text{-}3)$ & $93.35\,\textcolor{arrowdown}{\blacktriangledown}\;(\text{-}2)$ & $64.43\,\textcolor{arrowdown}{\blacktriangledown}\;(\text{-}2)$ & $93.41\,\textcolor{arrowdown}{\blacktriangledown}\;(\text{-}2)$ & $52.53\,\textcolor{arrowdown}{\blacktriangledown}\;(\text{-}1)$ \\
\bottomrule
\end{tabular}
\label{tab:0}
\end{table*}

\begin{table*}[t]
\centering
\footnotesize
\renewcommand{\arraystretch}{1.6}
\setlength{\tabcolsep}{2pt}
\definecolor{arrowup}{HTML}{007C97}
\definecolor{arrowdown}{HTML}{ED1C24}
\caption{Top-1 test accuracy (\%) across datasets and local learning rate $\log\eta$, with Dirichlet concentration $c=0.1$. 
Each cell shows the best result over $\log\Phi \in \{-1, -2, -3\}$ with the selected $\Phi$ value in parentheses. 
The best results per row are \underline{\textbf{bold and underlined}}. 
Here, {\textcolor{arrowup}{$\blacktriangle$}} indicates the best learning rate for each method within a dataset; {\textcolor{arrowdown}{$\blacktriangledown$}} otherwise.}
\begin{tabular}{cccccccccccc}
\toprule
 & $\log\eta$ & FedAvg & FedProx & FedDyn & SCAFFOLD & FedSAM & FedSpeed & FedSMOO & FedGamma & FedLESAM & FedWMSAM \\
\midrule
\multirow{3}{*}{\rotatebox{90}{Skin lesion}} & $\text{-}1$ & $14.29\,\textcolor{arrowdown}{\blacktriangledown}$ & $16.30\,\textcolor{arrowdown}{\blacktriangledown}\;(\text{-}3)$ & $16.29\,\textcolor{arrowdown}{\blacktriangledown}\;(\text{-}1)$ & $14.29\,\textcolor{arrowdown}{\blacktriangledown}$ & $14.29\,\textcolor{arrowdown}{\blacktriangledown}\;(\text{-}1)$ & $14.87\,\textcolor{arrowdown}{\blacktriangledown}\;(\text{-}3)$ & $14.39\,\textcolor{arrowdown}{\blacktriangledown}\;(\text{-}2)$ & $14.29\,\textcolor{arrowdown}{\blacktriangledown}\;(\text{-}1)$ & $14.29\,\textcolor{arrowdown}{\blacktriangledown}\;(\text{-}1)$ & $\underline{\mathbf{18.76\,\textcolor{arrowdown}{\blacktriangledown}\;(\text{-}2)}}$ \\
 & $\text{-}2$ & $30.10\,\textcolor{arrowdown}{\blacktriangledown}$ & $31.68\,\textcolor{arrowdown}{\blacktriangledown}\;(\text{-}3)$ & $19.46\,\textcolor{arrowdown}{\blacktriangledown}\;(\text{-}1)$ & $17.59\,\textcolor{arrowdown}{\blacktriangledown}$ & $\underline{\mathbf{47.91\,\textcolor{arrowdown}{\blacktriangledown}\;(\text{-}2)}}$ & $17.06\,\textcolor{arrowdown}{\blacktriangledown}\;(\text{-}1)$ & $20.82\,\textcolor{arrowdown}{\blacktriangledown}\;(\text{-}1)$ & $18.55\,\textcolor{arrowdown}{\blacktriangledown}\;(\text{-}2)$ & $26.86\,\textcolor{arrowdown}{\blacktriangledown}\;(\text{-}3)$ & $20.24\,\textcolor{arrowdown}{\blacktriangledown}\;(\text{-}3)$ \\
 & $\text{-}3$ & $66.12\,\textcolor{arrowup}{\blacktriangle}$ & $67.05\,\textcolor{arrowup}{\blacktriangle}\;(\text{-}2)$ & $\underline{\mathbf{69.60\,\textcolor{arrowup}{\blacktriangle}\;(\text{-}2)}}$ & $68.12\,\textcolor{arrowup}{\blacktriangle}$ & $68.57\,\textcolor{arrowup}{\blacktriangle}\;(\text{-}2)$ & $69.42\,\textcolor{arrowup}{\blacktriangle}\;(\text{-}2)$ & $69.53\,\textcolor{arrowup}{\blacktriangle}\;(\text{-}2)$ & $68.22\,\textcolor{arrowup}{\blacktriangle}\;(\text{-}3)$ & $66.63\,\textcolor{arrowup}{\blacktriangle}\;(\text{-}2)$ & $68.15\,\textcolor{arrowup}{\blacktriangle}\;(\text{-}3)$ \\
\midrule
\multirow{3}{*}{\rotatebox{90}{Blood cell}} & $\text{-}1$ & $12.50\,\textcolor{arrowdown}{\blacktriangledown}$ & $20.72\,\textcolor{arrowdown}{\blacktriangledown}\;(\text{-}2)$ & $13.08\,\textcolor{arrowdown}{\blacktriangledown}\;(\text{-}3)$ & $12.50\,\textcolor{arrowdown}{\blacktriangledown}$ & $12.50\,\textcolor{arrowdown}{\blacktriangledown}\;(\text{-}1)$ & $14.53\,\textcolor{arrowdown}{\blacktriangledown}\;(\text{-}3)$ & $12.91\,\textcolor{arrowdown}{\blacktriangledown}\;(\text{-}2)$ & $12.50\,\textcolor{arrowdown}{\blacktriangledown}\;(\text{-}1)$ & $16.60\,\textcolor{arrowdown}{\blacktriangledown}\;(\text{-}2)$ & $\underline{\mathbf{21.84\,\textcolor{arrowdown}{\blacktriangledown}\;(\text{-}2)}}$ \\
 & $\text{-}2$ & $67.44\,\textcolor{arrowdown}{\blacktriangledown}$ & $\underline{\mathbf{98.15\,\textcolor{arrowup}{\blacktriangle}\;(\text{-}2)}}$ & $52.32\,\textcolor{arrowdown}{\blacktriangledown}\;(\text{-}1)$ & $48.99\,\textcolor{arrowdown}{\blacktriangledown}$ & $66.60\,\textcolor{arrowdown}{\blacktriangledown}\;(\text{-}2)$ & $51.88\,\textcolor{arrowdown}{\blacktriangledown}\;(\text{-}1)$ & $68.60\,\textcolor{arrowdown}{\blacktriangledown}\;(\text{-}3)$ & $69.26\,\textcolor{arrowdown}{\blacktriangledown}\;(\text{-}3)$ & $97.17\,\textcolor{arrowdown}{\blacktriangledown}\;(\text{-}3)$ & $73.25\,\textcolor{arrowdown}{\blacktriangledown}\;(\text{-}3)$ \\
 & $\text{-}3$ & $97.44\,\textcolor{arrowup}{\blacktriangle}$ & $97.69\,\textcolor{arrowdown}{\blacktriangledown}\;(\text{-}3)$ & $97.81\,\textcolor{arrowup}{\blacktriangle}\;(\text{-}3)$ & $97.33\,\textcolor{arrowup}{\blacktriangle}$ & $\underline{\mathbf{98.24\,\textcolor{arrowup}{\blacktriangle}\;(\text{-}2)}}$ & $97.84\,\textcolor{arrowup}{\blacktriangle}\;(\text{-}2)$ & $97.87\,\textcolor{arrowup}{\blacktriangle}\;(\text{-}2)$ & $98.00\,\textcolor{arrowup}{\blacktriangle}\;(\text{-}2)$ & $97.69\,\textcolor{arrowup}{\blacktriangle}\;(\text{-}3)$ & $98.14\,\textcolor{arrowup}{\blacktriangle}\;(\text{-}3)$ \\
\midrule
\multirow{3}{*}{\rotatebox{90}{Colon path.}} & $\text{-}1$ & $11.11\,\textcolor{arrowdown}{\blacktriangledown}$ & $24.98\,\textcolor{arrowdown}{\blacktriangledown}\;(\text{-}2)$ & $12.24\,\textcolor{arrowdown}{\blacktriangledown}\;(\text{-}3)$ & $14.62\,\textcolor{arrowdown}{\blacktriangledown}$ & $20.12\,\textcolor{arrowdown}{\blacktriangledown}\;(\text{-}3)$ & $\underline{\mathbf{27.39\,\textcolor{arrowdown}{\blacktriangledown}\;(\text{-}1)}}$ & $11.17\,\textcolor{arrowdown}{\blacktriangledown}\;(\text{-}3)$ & $12.81\,\textcolor{arrowdown}{\blacktriangledown}\;(\text{-}1)$ & $11.11\,\textcolor{arrowdown}{\blacktriangledown}\;(\text{-}1)$ & $24.46\,\textcolor{arrowdown}{\blacktriangledown}\;(\text{-}3)$ \\
 & $\text{-}2$ & $84.39\,\textcolor{arrowdown}{\blacktriangledown}$ & $83.93\,\textcolor{arrowdown}{\blacktriangledown}\;(\text{-}2)$ & $45.69\,\textcolor{arrowdown}{\blacktriangledown}\;(\text{-}1)$ & $16.96\,\textcolor{arrowdown}{\blacktriangledown}$ & $\underline{\mathbf{88.11\,\textcolor{arrowdown}{\blacktriangledown}\;(\text{-}2)}}$ & $69.32\,\textcolor{arrowdown}{\blacktriangledown}\;(\text{-}1)$ & $69.91\,\textcolor{arrowdown}{\blacktriangledown}\;(\text{-}1)$ & $21.96\,\textcolor{arrowdown}{\blacktriangledown}\;(\text{-}3)$ & $82.55\,\textcolor{arrowdown}{\blacktriangledown}\;(\text{-}3)$ & $27.90\,\textcolor{arrowdown}{\blacktriangledown}\;(\text{-}2)$ \\
 & $\text{-}3$ & $92.86\,\textcolor{arrowup}{\blacktriangle}$ & $93.52\,\textcolor{arrowup}{\blacktriangle}\;(\text{-}2)$ & $93.66\,\textcolor{arrowup}{\blacktriangle}\;(\text{-}3)$ & $90.88\,\textcolor{arrowup}{\blacktriangle}$ & $\underline{\mathbf{94.31\,\textcolor{arrowup}{\blacktriangle}\;(\text{-}1)}}$ & $93.55\,\textcolor{arrowup}{\blacktriangle}\;(\text{-}3)$ & $93.59\,\textcolor{arrowup}{\blacktriangle}\;(\text{-}2)$ & $93.04\,\textcolor{arrowup}{\blacktriangle}\;(\text{-}2)$ & $93.47\,\textcolor{arrowup}{\blacktriangle}\;(\text{-}2)$ & $78.59\,\textcolor{arrowup}{\blacktriangle}\;(\text{-}3)$ \\
\bottomrule
\end{tabular}
\label{tab:1}
\end{table*}

\begin{figure*}[t]
\centering
\includegraphics[width=\textwidth]{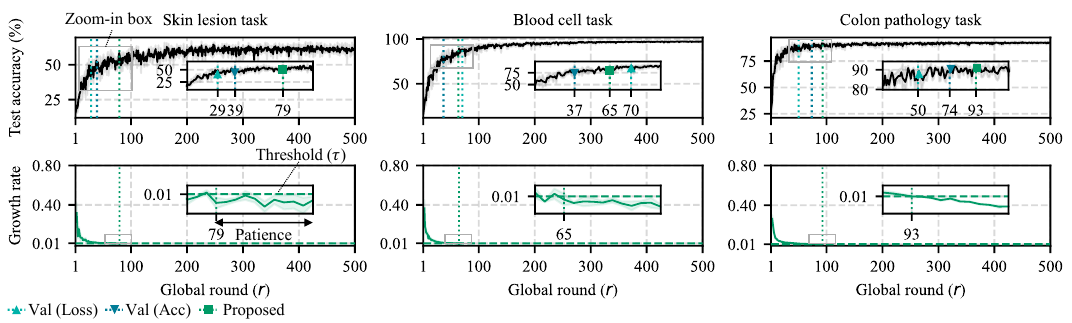}
\caption{Test accuracy (\%) and growth rate trajectories over global rounds with $\tau=0.01$ and $\rho=10$ for both proposed and validation-based early stopping. The zoom-in boxes highlight the regions around the early stopping points.}
\label{fig:result1}
\end{figure*}

\begin{table*}[t]
\centering
\caption{Performance comparison under diverse non-IID distributions and coefficient values $c$ with $\tau=0.01$ and $\rho=10$. Each value denotes the mean test accuracy difference between the proposed stopping point and the best-performing validation-based stopping point. The positive and negative values are highlighted in \colorbox{myteal!15}{blue} and \colorbox{myred!15}{red}, respectively, where color intensity reflects the absolute magnitude of the difference. Note that \underline{\textbf{bold and underlined}} values indicate the largest value for each $c$.}
\label{tab:2}
\resizebox{\textwidth}{!}{
\fontsize{6.0pt}{6.0pt}\selectfont
\begin{tabular}{c lccccccccc}
\toprule
 & \multirow{2}{*}{\textbf{Method}} & \multicolumn{3}{c}{\emph{Label skew} (Dirichlet)} & \multicolumn{3}{c}{\emph{Label skew} (Pathological)} & \multicolumn{3}{c}{\emph{Quantity skew}} \\
\cmidrule(lr){3-5} \cmidrule(lr){6-8} \cmidrule(lr){9-11}
 &  & $c=0.01$ & $c=0.1$ & $c=1.0$ & $c=1$ & $c=2$ & $c=3$ & $c=0.01$ & $c=0.1$ & $c=1.0$ \\
\midrule
\multirow{10}{*}{\rotatebox[origin=c]{90}{Skin lesion task}} & FedAvg & \cellcolor{myteal!44}{\textcolor{black}{+16.55}} & \cellcolor{myteal!9}{\textcolor{black}{+3.42}} & \cellcolor{myteal!5}{\textcolor{black}{+0.68}} & \cellcolor{myteal!8}{\textcolor{black}{+3.02}} & \cellcolor{myteal!42}{\textcolor{black}{+15.84}} & \cellcolor{myred!8}{\textcolor{black}{-2.98}} & \cellcolor{myteal!28}{\textcolor{black}{+10.61}} & \cellcolor{myteal!10}{\textcolor{black}{+4.08}} & \cellcolor{myred!5}{\textcolor{black}{-0.26}} \\
 & FedProx & \cellcolor{myteal!33}{\textcolor{black}{+12.62}} & \cellcolor{myteal!39}{\textcolor{black}{+14.52}} & \cellcolor{myteal!7}{\textcolor{black}{+2.93}} & \cellcolor{myteal!31}{\textcolor{black}{+11.74}} & \cellcolor{myteal!28}{\textcolor{black}{+10.50}} & \cellcolor{myteal!5}{\textcolor{black}{+1.93}} & \cellcolor{myteal!28}{\textcolor{black}{+10.47}} & \cellcolor{myteal!20}{\textcolor{black}{+7.46}} & \cellcolor{myteal!14}{\textcolor{black}{+5.29}} \\
 & FedDyn & \cellcolor{myteal!56}{\textcolor{black}{+20.87}} & \cellcolor{myteal!41}{\textcolor{black}{+15.49}} & \cellcolor{myred!6}{\textcolor{black}{-2.40}} & \cellcolor{myteal!17}{\textcolor{black}{+6.56}} & \cellcolor{myteal!35}{\textcolor{black}{+13.06}} & \cellcolor{myteal!12}{\textcolor{black}{+4.66}} & \cellcolor{myteal!48}{\textcolor{black}{+18.07}} & \cellcolor{myteal!12}{\textcolor{black}{+4.78}} & \cellcolor{myteal!23}{\textcolor{black}{+8.78}} \\
 & SCAFFOLD & \cellcolor{myteal!41}{\textcolor{black}{+15.53}} & \cellcolor{myteal!5}{\textcolor{black}{+1.92}} & \cellcolor{myteal!17}{\textcolor{black}{+6.55}} & \cellcolor{myteal!20}{\textcolor{black}{+7.53}} & \cellcolor{myteal!8}{\textcolor{black}{+3.18}} & \cellcolor{myteal!16}{\textcolor{black}{+6.11}} & \cellcolor{myred!5}{\textcolor{black}{-1.73}} & \cellcolor{myteal!8}{\textcolor{black}{+3.07}} & \cellcolor{myteal!22}{\textcolor{black}{+8.45}} \\
 & FedSAM & \cellcolor{myteal!32}{\textcolor{black}{+12.18}} & \cellcolor{myteal!36}{\textcolor{black}{+13.44}} & \cellcolor{myred!5}{\textcolor{black}{-0.15}} & \cellcolor{myteal!6}{\textcolor{black}{+2.50}} & \cellcolor{myteal!5}{\textcolor{black}{+1.80}} & \cellcolor{myteal!9}{\textcolor{black}{+3.56}} & \cellcolor{myteal!28}{\textcolor{black}{+10.53}} & \cellcolor{myteal!15}{\textcolor{black}{+5.67}} & \cellcolor{myteal!20}{\textcolor{black}{+7.58}} \\
 & FedSpeed & \cellcolor{myteal!66}{\textcolor{white}{\textbf{\underline{+24.82}}}} & \cellcolor{myteal!44}{\textcolor{black}{+16.51}} & \cellcolor{myteal!5}{\textcolor{black}{+0.60}} & \cellcolor{myteal!69}{\textcolor{white}{\textbf{\underline{+25.82}}}} & \cellcolor{myteal!45}{\textcolor{black}{+16.94}} & \cellcolor{myteal!5}{\textcolor{black}{+1.45}} & \cellcolor{myteal!67}{\textcolor{white}{+24.94}} & \cellcolor{myteal!20}{\textcolor{black}{\textbf{\underline{+7.62}}}} & \cellcolor{myteal!25}{\textcolor{black}{+9.43}} \\
 & FedSMOO & \cellcolor{myteal!50}{\textcolor{black}{+18.85}} & \cellcolor{myteal!47}{\textcolor{black}{+17.60}} & \cellcolor{myteal!5}{\textcolor{black}{+0.81}} & \cellcolor{myteal!64}{\textcolor{white}{+24.06}} & \cellcolor{myteal!22}{\textcolor{black}{+8.51}} & \cellcolor{myteal!6}{\textcolor{black}{+2.34}} & \cellcolor{myteal!73}{\textcolor{white}{\textbf{\underline{+27.17}}}} & \cellcolor{myteal!13}{\textcolor{black}{+5.06}} & \cellcolor{myteal!33}{\textcolor{black}{\textbf{\underline{+12.57}}}} \\
 & FedGamma & \cellcolor{myteal!5}{\textcolor{black}{+1.07}} & \cellcolor{myteal!12}{\textcolor{black}{+4.64}} & \cellcolor{myteal!8}{\textcolor{black}{+3.33}} & \cellcolor{myteal!7}{\textcolor{black}{+2.76}} & \cellcolor{myteal!7}{\textcolor{black}{+2.76}} & \cellcolor{myteal!10}{\textcolor{black}{+3.78}} & \cellcolor{myteal!5}{\textcolor{black}{+0.54}} & \cellcolor{myteal!5}{\textcolor{black}{+0.78}} & \cellcolor{myteal!21}{\textcolor{black}{+7.90}} \\
 & FedLESAM & \cellcolor{myteal!52}{\textcolor{black}{+19.65}} & \cellcolor{myteal!13}{\textcolor{black}{+5.17}} & \cellcolor{myteal!5}{\textcolor{black}{+1.36}} & \cellcolor{myred!14}{\textcolor{black}{-5.38}} & \cellcolor{myteal!9}{\textcolor{black}{+3.38}} & \cellcolor{myteal!12}{\textcolor{black}{+4.74}} & \cellcolor{myteal!40}{\textcolor{black}{+15.15}} & \cellcolor{myteal!18}{\textcolor{black}{+6.97}} & \cellcolor{myteal!10}{\textcolor{black}{+3.98}} \\
 & FedWMSAM & \cellcolor{myteal!42}{\textcolor{black}{+15.73}} & \cellcolor{myteal!51}{\textcolor{black}{\textbf{\underline{+19.14}}}} & \cellcolor{myteal!47}{\textcolor{black}{\textbf{\underline{+17.81}}}} & \cellcolor{myteal!52}{\textcolor{black}{+19.36}} & \cellcolor{myteal!79}{\textcolor{white}{\textbf{\underline{+29.59}}}} & \cellcolor{myteal!46}{\textcolor{black}{\textbf{\underline{+17.30}}}} & \cellcolor{myred!5}{\textcolor{black}{-0.23}} & \cellcolor{myred!5}{\textcolor{black}{-1.33}} & \cellcolor{myteal!10}{\textcolor{black}{+3.78}} \\
\midrule
\multirow{10}{*}{\rotatebox[origin=c]{90}{Blood cell task}} & FedAvg & \cellcolor{myteal!20}{\textcolor{black}{+7.69}} & \cellcolor{myred!5}{\textcolor{black}{-0.15}} & \cellcolor{myred!5}{\textcolor{black}{-1.34}} & \cellcolor{myred!32}{\textcolor{black}{-12.08}} & \cellcolor{myred!5}{\textcolor{black}{-1.26}} & \cellcolor{myteal!8}{\textcolor{black}{\textbf{\underline{+3.06}}}} & \cellcolor{myred!5}{\textcolor{black}{-0.23}} & \cellcolor{myred!5}{\textcolor{black}{-0.53}} & \cellcolor{myred!5}{\textcolor{black}{-0.40}} \\
 & FedProx & \cellcolor{myteal!14}{\textcolor{black}{+5.33}} & \cellcolor{myred!5}{\textcolor{black}{-0.20}} & \cellcolor{myred!5}{\textcolor{black}{-0.95}} & \cellcolor{myteal!11}{\textcolor{black}{+4.37}} & \cellcolor{myteal!5}{\textcolor{black}{+0.09}} & \cellcolor{myteal!5}{\textcolor{black}{+0.94}} & \cellcolor{myteal!5}{\textcolor{black}{+0.55}} & \cellcolor{myteal!5}{\textcolor{black}{+0.06}} & \cellcolor{myred!5}{\textcolor{black}{-1.79}} \\
 & FedDyn & \cellcolor{myteal!66}{\textcolor{white}{\textbf{\underline{+24.88}}}} & \cellcolor{myteal!14}{\textcolor{black}{+5.49}} & \cellcolor{myteal!5}{\textcolor{black}{+0.00}} & \cellcolor{myteal!74}{\textcolor{white}{+27.70}} & \cellcolor{myteal!21}{\textcolor{black}{+8.08}} & \cellcolor{myteal!5}{\textcolor{black}{+1.90}} & \cellcolor{myteal!5}{\textcolor{black}{+0.27}} & \cellcolor{myteal!5}{\textcolor{black}{+0.37}} & \cellcolor{myteal!5}{\textcolor{black}{+0.40}} \\
 & SCAFFOLD & \cellcolor{myred!11}{\textcolor{black}{-4.14}} & \cellcolor{myred!5}{\textcolor{black}{-1.18}} & \cellcolor{myteal!5}{\textcolor{black}{\textbf{\underline{+1.31}}}} & \cellcolor{myteal!12}{\textcolor{black}{+4.49}} & \cellcolor{myteal!5}{\textcolor{black}{+0.83}} & \cellcolor{myteal!5}{\textcolor{black}{+0.82}} & \cellcolor{myteal!5}{\textcolor{black}{+1.08}} & \cellcolor{myteal!5}{\textcolor{black}{\textbf{\underline{+0.61}}}} & \cellcolor{myteal!9}{\textcolor{black}{\textbf{\underline{+3.42}}}} \\
 & FedSAM & \cellcolor{myteal!21}{\textcolor{black}{+7.91}} & \cellcolor{myteal!9}{\textcolor{black}{+3.59}} & \cellcolor{myred!5}{\textcolor{black}{-1.21}} & \cellcolor{myteal!30}{\textcolor{black}{+11.23}} & \cellcolor{myteal!9}{\textcolor{black}{+3.71}} & \cellcolor{myteal!5}{\textcolor{black}{+1.25}} & \cellcolor{myteal!5}{\textcolor{black}{+0.85}} & \cellcolor{myteal!5}{\textcolor{black}{+0.37}} & \cellcolor{myred!5}{\textcolor{black}{-0.32}} \\
 & FedSpeed & \cellcolor{myteal!37}{\textcolor{black}{+13.96}} & \cellcolor{myteal!13}{\textcolor{black}{+4.90}} & \cellcolor{myteal!5}{\textcolor{black}{+0.06}} & \cellcolor{myteal!100}{\textcolor{white}{\textbf{\underline{+37.16}}}} & \cellcolor{myteal!22}{\textcolor{black}{+8.50}} & \cellcolor{myteal!5}{\textcolor{black}{+1.10}} & \cellcolor{myred!5}{\textcolor{black}{-0.10}} & \cellcolor{myred!5}{\textcolor{black}{-0.12}} & \cellcolor{myteal!5}{\textcolor{black}{+0.26}} \\
 & FedSMOO & \cellcolor{myteal!43}{\textcolor{black}{+16.08}} & \cellcolor{myteal!12}{\textcolor{black}{+4.71}} & \cellcolor{myteal!5}{\textcolor{black}{+0.56}} & \cellcolor{myteal!99}{\textcolor{white}{+37.14}} & \cellcolor{myteal!23}{\textcolor{black}{\textbf{\underline{+8.60}}}} & \cellcolor{myteal!5}{\textcolor{black}{+0.83}} & \cellcolor{myred!5}{\textcolor{black}{-0.05}} & \cellcolor{myred!5}{\textcolor{black}{-0.07}} & \cellcolor{myteal!5}{\textcolor{black}{+0.39}} \\
 & FedGamma & \cellcolor{myteal!7}{\textcolor{black}{+2.93}} & \cellcolor{myred!5}{\textcolor{black}{-0.56}} & \cellcolor{myteal!5}{\textcolor{black}{+0.51}} & \cellcolor{myteal!17}{\textcolor{black}{+6.53}} & \cellcolor{myred!5}{\textcolor{black}{-1.52}} & \cellcolor{myteal!5}{\textcolor{black}{+0.03}} & \cellcolor{myteal!16}{\textcolor{black}{\textbf{\underline{+6.08}}}} & \cellcolor{myred!13}{\textcolor{black}{-4.88}} & \cellcolor{myteal!8}{\textcolor{black}{+3.18}} \\
 & FedLESAM & \cellcolor{myred!5}{\textcolor{black}{-0.29}} & \cellcolor{myteal!7}{\textcolor{black}{+2.77}} & \cellcolor{myred!5}{\textcolor{black}{-1.01}} & \cellcolor{myteal!19}{\textcolor{black}{+7.42}} & \cellcolor{myteal!13}{\textcolor{black}{+5.20}} & \cellcolor{myred!5}{\textcolor{black}{-1.13}} & \cellcolor{myred!5}{\textcolor{black}{-0.15}} & \cellcolor{myteal!5}{\textcolor{black}{+0.17}} & \cellcolor{myred!5}{\textcolor{black}{-1.32}} \\
 & FedWMSAM & \cellcolor{myred!5}{\textcolor{black}{-1.73}} & \cellcolor{myteal!42}{\textcolor{black}{\textbf{\underline{+15.77}}}} & \cellcolor{myred!15}{\textcolor{black}{-5.82}} & \cellcolor{myteal!45}{\textcolor{black}{+16.89}} & \cellcolor{myred!19}{\textcolor{black}{-7.32}} & \cellcolor{myred!14}{\textcolor{black}{-5.33}} & \cellcolor{myred!11}{\textcolor{black}{-4.43}} & \cellcolor{myred!5}{\textcolor{black}{-0.22}} & \cellcolor{myred!34}{\textcolor{black}{-12.92}} \\
\midrule
\multirow{10}{*}{\rotatebox[origin=c]{90}{Colon pathology task}} & FedAvg & \cellcolor{myteal!28}{\textcolor{black}{+10.54}} & \cellcolor{myteal!7}{\textcolor{black}{+2.88}} & \cellcolor{myteal!5}{\textcolor{black}{+0.39}} & \cellcolor{myteal!5}{\textcolor{black}{+1.90}} & \cellcolor{myteal!25}{\textcolor{black}{+9.65}} & \cellcolor{myteal!5}{\textcolor{black}{+0.20}} & \cellcolor{myred!5}{\textcolor{black}{-0.86}} & \cellcolor{myred!5}{\textcolor{black}{-0.09}} & \cellcolor{myred!5}{\textcolor{black}{-0.21}} \\
 & FedProx & \cellcolor{myteal!57}{\textcolor{black}{+21.27}} & \cellcolor{myteal!9}{\textcolor{black}{+3.53}} & \cellcolor{myred!5}{\textcolor{black}{-0.03}} & \cellcolor{myteal!48}{\textcolor{black}{+17.98}} & \cellcolor{myteal!37}{\textcolor{black}{+14.01}} & \cellcolor{myteal!18}{\textcolor{black}{+6.69}} & \cellcolor{myred!5}{\textcolor{black}{-1.17}} & \cellcolor{myred!5}{\textcolor{black}{-0.43}} & \cellcolor{myred!5}{\textcolor{black}{-0.12}} \\
 & FedDyn & \cellcolor{myteal!35}{\textcolor{black}{+13.37}} & \cellcolor{myteal!5}{\textcolor{black}{+1.53}} & \cellcolor{myred!5}{\textcolor{black}{-1.19}} & \cellcolor{myred!5}{\textcolor{black}{-1.56}} & \cellcolor{myteal!17}{\textcolor{black}{+6.42}} & \cellcolor{myteal!16}{\textcolor{black}{+5.95}} & \cellcolor{myred!5}{\textcolor{black}{-1.02}} & \cellcolor{myred!6}{\textcolor{black}{-2.58}} & \cellcolor{myred!5}{\textcolor{black}{-1.58}} \\
 & SCAFFOLD & \cellcolor{myteal!5}{\textcolor{black}{+1.67}} & \cellcolor{myteal!18}{\textcolor{black}{+6.90}} & \cellcolor{myteal!14}{\textcolor{black}{\textbf{\underline{+5.22}}}} & \cellcolor{myred!15}{\textcolor{black}{-5.67}} & \cellcolor{myteal!10}{\textcolor{black}{+3.97}} & \cellcolor{myteal!5}{\textcolor{black}{+1.90}} & \cellcolor{myteal!7}{\textcolor{black}{+2.62}} & \cellcolor{myteal!29}{\textcolor{black}{\textbf{\underline{+10.93}}}} & \cellcolor{myred!5}{\textcolor{black}{-0.89}} \\
 & FedSAM & \cellcolor{myteal!22}{\textcolor{black}{+8.50}} & \cellcolor{myteal!5}{\textcolor{black}{+1.37}} & \cellcolor{myred!5}{\textcolor{black}{-0.53}} & \cellcolor{myteal!65}{\textcolor{white}{+24.39}} & \cellcolor{myteal!28}{\textcolor{black}{+10.43}} & \cellcolor{myteal!7}{\textcolor{black}{+2.60}} & \cellcolor{myred!5}{\textcolor{black}{-1.42}} & \cellcolor{myred!5}{\textcolor{black}{-0.38}} & \cellcolor{myred!5}{\textcolor{black}{\textbf{\underline{-0.11}}}} \\
 & FedSpeed & \cellcolor{myteal!62}{\textcolor{white}{+23.06}} & \cellcolor{myteal!5}{\textcolor{black}{+0.61}} & \cellcolor{myred!5}{\textcolor{black}{-1.15}} & \cellcolor{myred!6}{\textcolor{black}{-2.23}} & \cellcolor{myteal!21}{\textcolor{black}{+7.90}} & \cellcolor{myteal!17}{\textcolor{black}{+6.42}} & \cellcolor{myred!5}{\textcolor{black}{-0.49}} & \cellcolor{myred!5}{\textcolor{black}{-1.74}} & \cellcolor{myred!5}{\textcolor{black}{-1.01}} \\
 & FedSMOO & \cellcolor{myteal!85}{\textcolor{white}{\textbf{\underline{+31.70}}}} & \cellcolor{myteal!5}{\textcolor{black}{+0.01}} & \cellcolor{myred!8}{\textcolor{black}{-3.30}} & \cellcolor{myteal!51}{\textcolor{black}{+19.04}} & \cellcolor{myteal!45}{\textcolor{black}{\textbf{\underline{+16.78}}}} & \cellcolor{myteal!18}{\textcolor{black}{+6.93}} & \cellcolor{myred!5}{\textcolor{black}{-0.32}} & \cellcolor{myred!5}{\textcolor{black}{-1.66}} & \cellcolor{myred!5}{\textcolor{black}{-1.29}} \\
 & FedGamma & \cellcolor{myred!8}{\textcolor{black}{-3.32}} & \cellcolor{myteal!12}{\textcolor{black}{+4.59}} & \cellcolor{myteal!9}{\textcolor{black}{+3.43}} & 0.00 & \cellcolor{myteal!17}{\textcolor{black}{+6.32}} & \cellcolor{myteal!14}{\textcolor{black}{+5.47}} & \cellcolor{myteal!13}{\textcolor{black}{\textbf{\underline{+5.08}}}} & \cellcolor{myteal!7}{\textcolor{black}{+2.71}} & \cellcolor{myred!5}{\textcolor{black}{-1.81}} \\
 & FedLESAM & \cellcolor{myteal!10}{\textcolor{black}{+4.04}} & \cellcolor{myteal!30}{\textcolor{black}{+11.41}} & \cellcolor{myred!5}{\textcolor{black}{-1.07}} & \cellcolor{myteal!96}{\textcolor{white}{\textbf{\underline{+35.86}}}} & \cellcolor{myteal!42}{\textcolor{black}{+15.78}} & \cellcolor{myteal!10}{\textcolor{black}{+3.77}} & \cellcolor{myred!5}{\textcolor{black}{-0.75}} & \cellcolor{myteal!5}{\textcolor{black}{+0.03}} & \cellcolor{myred!5}{\textcolor{black}{-0.37}} \\
 & FedWMSAM & \cellcolor{myteal!60}{\textcolor{black}{+22.44}} & \cellcolor{myteal!66}{\textcolor{white}{\textbf{\underline{+24.68}}}} & \cellcolor{myred!17}{\textcolor{black}{-6.53}} & \cellcolor{myteal!10}{\textcolor{black}{+3.96}} & \cellcolor{myteal!44}{\textcolor{black}{+16.54}} & \cellcolor{myteal!34}{\textcolor{black}{\textbf{\underline{+12.86}}}} & \cellcolor{myteal!5}{\textcolor{black}{+0.34}} & \cellcolor{myred!7}{\textcolor{black}{-2.87}} & \cellcolor{myred!15}{\textcolor{black}{-5.88}} \\
\bottomrule
\end{tabular}
}
\end{table*}

\begin{figure*}[t]
\centering
\includegraphics[width=\textwidth]{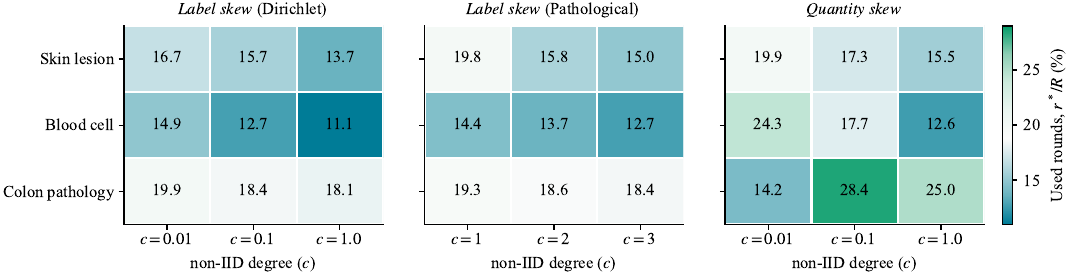}
\caption{ Used global round ratio $r^*/R$ (\%) of FedAvg across non-IID degrees $c$ for different datasets and \emph{data skew} settings. 
Each cell reports the average used-round ratio, with smaller and larger values denoted by \textbf{\textcolor{myteal}{blue}} and \textbf{\textcolor{mygreen}{green}}, respectively. 
}
\label{fig:result2}
\end{figure*}

\begin{figure*}[t]
\centering
\includegraphics[width=\textwidth]{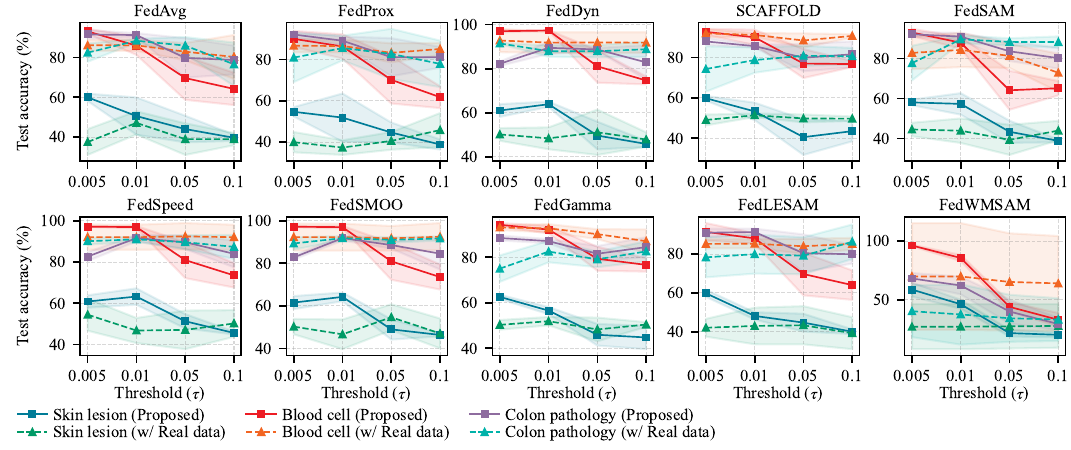}
\caption{Test accuracy (\%) of FL methods evaluated under various $\tau$ values with $\rho=10$ for both validation-based and proposed early stopping.
The solid curves denote the mean test accuracy, while the shaded regions indicate the standard deviation.
}
\label{fig:result3}
\end{figure*}

\begin{table*}[t]
\centering
\footnotesize
\renewcommand{\arraystretch}{1.2}
\setlength{\tabcolsep}{3.5pt}
\caption{Performance and stopping behavior of the proposed approach under $\tau=0.1$ and $\rho=10$, under \emph{label skew} (Dirichlet) with $c=0.1$. 
The $\Delta_{Acc.}$ and $\Delta_{r}$ denote the accuracy gain and round difference, respectively, relative to the best validation-based baseline. 
Note that the $r^*/R$ (\%) denotes the percentage of the fixed-round budget used, where $R=500$.}
\label{tab:3}
\begin{tabular}{l*{15}{c}}
\toprule
\multirow{2}{*}{\textbf{Method}} & \multicolumn{5}{c}{Skin lesion task} & \multicolumn{5}{c}{Blood cell task} & \multicolumn{5}{c}{Colon pathology task} \\
\cmidrule(lr){2-6} \cmidrule(lr){7-11} \cmidrule(lr){12-16}
 & \textit{Acc.} (\%) & $r^*$ & $\Delta_{Acc.}$ & $\Delta_{r}$ & $r^*/R$ (\%) & \textit{Acc.} (\%) & $r^*$ & $\Delta_{Acc.}$ & $\Delta_{r}$ & $r^*/R$ (\%) & \textit{Acc.} (\%) & $r^*$ & $\Delta_{Acc.}$ & $\Delta_{r}$ & $r^*/R$ (\%) \\
\midrule
FedAvg & 14.29 & 15.0 & - & +4.7 & 3.0 & 12.50 & 16.0 & - & +6.0 & 3.2 & 11.11 & 18.0 & - & +8.0 & 3.6 \\
FedProx & 14.29 & 14.7 & - & +4.7 & 2.9 & 12.50 & 15.0 & - & +4.7 & 3.0 & 11.11 & 18.3 & - & +8.3 & 3.7 \\
FedDyn & 14.29 & 19.0 & - & +9.0 & 3.8 & 12.50 & 19.0 & - & +9.0 & 3.8 & 11.11 & 21.3 & - & +11.3 & 4.3 \\
SCAFFOLD & 14.29 & 26.0 & - & +16.0 & 5.2 & 12.50 & 22.3 & - & +12.3 & 4.5 & 11.11 & 43.7 & - & +33.7 & 8.7 \\
FedSAM & 14.29 & 15.0 & - & +5.0 & 3.0 & 12.50 & 14.0 & - & +4.0 & 2.8 & 11.11 & 18.0 & - & +8.0 & 3.6 \\
FedSpeed & 14.29 & 18.0 & - & +8.0 & 3.6 & 12.50 & 19.3 & - & +9.3 & 3.9 & 11.11 & 21.0 & - & +11.0 & 4.2 \\
FedSMOO & 14.29 & 17.7 & - & +7.7 & 3.5 & 12.50 & 17.7 & - & +7.7 & 3.5 & 11.11 & 21.3 & - & +11.3 & 4.3 \\
FedGamma & 14.29 & 25.3 & - & +15.3 & 5.1 & 12.50 & 24.0 & - & +14.0 & 4.8 & 11.11 & 43.3 & - & +33.3 & 8.7 \\
FedLESAM & 14.29 & 15.0 & - & +5.0 & 3.0 & 12.50 & 15.7 & - & +5.7 & 3.1 & 11.11 & 18.0 & - & +8.0 & 3.6 \\
FedWMSAM & 14.29 & 21.0 & -0.18 & +10.7 & 4.2 & 12.50 & 21.0 & - & +8.3 & 4.2 & 11.11 & 21.0 & -1.76 & +11.0 & 4.2 \\
\midrule
\multicolumn{16}{@{}l}{\textit{Note.} The symbol `-' denotes no difference.} \\
\end{tabular}
\end{table*}

\section{Experiment and Results}\label{sec:experiment}
\subsection{Experiment Setting}
We evaluate the proposed approach against validation-based early stopping using validation loss or accuracy on skin lesion \cite{tschandl2018ham10000}, blood cell \cite{acevedo2020dataset}, and colon pathology \cite{kather2019predicting} image classification tasks.
Note that the data-driven early stopping uses both training and validation splits in~\cite{medmnistv2}, while our approach uses only the training subset.
All clients employ ConvNeXtV2 \cite{woo2023convnext} as the local AI model.
We benchmark with recent FL methods, including FedAvg, FedProx, SCAFFOLD, FedDyn, FedSAM, FedSpeed, FedSMOO, FedGamma, FedLESAM, and FedWMSAM.
To simulate the federated setting, the dataset is partitioned across $N=100$ clients, and a subset of $M=10$ clients is randomly sampled for local training at each round.
The experiments were repeated with $3$ random seeds and run on AMD MI300X AI accelerators \cite{10930746}.

\subsection{Numerical Results}
\subsubsection{Sensitivity to FL Configurations}
To show how \emph{data skew} and the key hyperparameter $\Phi$ affect performance, Table~\ref{tab:0} reports the Top-1 accuracy across diverse FL configurations and datasets.
Note that neither FedAvg nor SCAFFOLD requires $\Phi$ on the client or server side.
For the two \emph{label skew} settings, FedSAM achieves the highest dataset-averaged accuracy, with 87.0\% under \emph{label skew} (Dirichlet, $L_d$) and 87.7\% under \emph{label skew} (Pathological, $L_p$).
Under \emph{Quantity skew} ($Q$), however, some recent FL methods underperform FedAvg, as FedGamma and FedWMSAM obtain dataset-averaged accuracies of only 41.0\% and 32.9\%, respectively, compared with FedAvg at 82.4\%.
Notably, $\Phi$ also shows no clear pattern, as the optimal value for performance varies across FL methods and \emph{data skew}.
Thus, adopting a recent FL method does not guarantee higher accuracy, as performance is affected by the \emph{data skew} and hyperparameter settings.

\subsubsection{Sensitivity to Client-side Configuration}
To verify that FL performance is also affected by client-side settings, Table~\ref{tab:1} reports the Top-1 accuracy under \emph{label skew} (Dirichlet) with $c=0.1$.
Here, both the client-side learning rate $\eta \in \{0.001, 0.01, 0.1\}$ and the FL key hyperparameter $\Phi$ are varied, while all other configurations are fixed for all datasets and FL methods.
For the skin lesion task, the FL method-averaged accuracy increases from $15.2\%$ at $\eta=0.1$ to $25.0\%$ at $\eta=0.01$ and $68.1\%$ at $\eta=0.001$.
Similarly, the blood cell task improves from $15.0\%$ to $69.4\%$ and $97.8\%$ as $\eta$ decreases.
The colon pathology task shows the same tendency, increasing from $17.0\%$ to $59.1\%$ and $91.7\%$.
These results indicate that FL performance is jointly shaped by client-side configuration, \emph{data skew}, the selected FL method, and the FL key hyperparameter $\Phi$.
Note that the FL method-averaged accuracy generally improves as $\eta$ decreases, motivating the use of $\eta=0.001$ in the following experiments.

\subsubsection{Effectiveness of Proposed Framework}
To show the effectiveness of the proposed framework, as shown in Fig.~\ref{fig:result1}, we evaluate FedAvg under \emph{label skew} (Dirichlet) with $c=0.1$ and compare it against validation-based early stopping.
On the skin lesion task, our approach stops, on average, $+45$ rounds later than validation-based early stopping, while achieving more than $+12.3\%$ higher performance.
In particular, the loss- and accuracy-based validation criteria stop at $r=29$ and $r=39$ with $44.86\%$ and $47.17\%$, whereas the proposed metric continues to round $79$ and reaches $58.34\%$.
For the blood cell task, training is extended by $+12$ rounds on average, yielding a mean performance gain of $+8.9\%$.

On the colon pathology task, the proposed criterion stops $+31$ rounds later than validation-based early stopping, yielding a mean performance gain of $+3.9\%$.
Specifically, the validation criteria stop at $r=50$ and $r=74$ with $84.53\%$ and $88.58\%$, while the proposed metric extends to $r=93$ and achieves the highest accuracy of $90.45\%$.
Moreover, the lower panels of Fig.~\ref{fig:result1} show that the proposed growth rate metric gradually decays over rounds, leading to stable stopping at the predefined threshold $\tau=0.01$.
Overall, the results indicate that the proposed early stopping framework achieves validation-level performance without requiring any validation data.

\subsubsection{Efficiency of Proposed Framework}
To show the efficiency of the proposed framework, as shown in Fig.~\ref{fig:result2}, we report the used global round ratio $r^*/R$ (\%) of FedAvg across non-IID degrees $c$ for different datasets and \emph{data skew} settings.
Here, we set $\tau=0.01$ and $\rho=10$ for all reported settings.
Across all considered settings, the proposed framework consumes at most $28.4\%$ of the fixed-round budget ($R=500$), confirming that early stopping substantially reduces computational resource waste.
For both \emph{label skew} settings, the used round ratio tends to decrease as $c$ increases, since smaller skew accelerates the saturation of the task vector trajectory.
In particular, under \emph{label skew} (Dirichlet), the blood cell task drops from $14.9\%$ at $c=0.01$ to $11.1\%$ at $c=1.0$, while the skin lesion task drops from $16.7\%$ to $13.7\%$.

Under \emph{quantity skew}, the proposed approach remains efficient but shows a less monotonic trend across $c$ values.
In detail, the colon pathology task uses $14.2\%$ at $c=0.01$, peaks at $28.4\%$ at $c=0.1$, and settles at $25.0\%$ at $c=1.0$.
Note that the blood cell task consumes the fewest rounds on average at $14.9\%$, whereas the colon pathology task consumes the most at $20.0\%$.
Thus, the results indicate that the proposed early stopping framework stops training within roughly $11\%$--$28\%$ of the fixed-round budget, achieving practical efficiency.

\subsubsection{Impact of non-IID Data Distributions}
To analyze the impact of non-IID data distributions, we evaluate the proposed framework under three representative data skew types across multiple $c$ values, as shown in Table \ref{tab:2}.
In detail, we analyze the performance differences at the respective stopping points between the proposed and validation-based early stopping approach.
For the skin lesion task, the proposed approach achieves large average gains under severe heterogeneity ($c=0.01/1$), reaching approximately $+15.8\%$, $+9.8\%$, and $+11.6\%$ for \emph{label skew} (Dirichlet, Pathological), and \emph{quantity skew}, respectively.
As $c$ increases to $1.0/3$, the average gains decrease to around $+3.2\%$, $+4.3\%$, and $+6.8\%$, indicating natural alignment with validation-based stopping as data distributions become less heterogeneous.

For the blood cell task, a similar trend is observed, where the proposed framework yields average gains of approximately $+7.3\%$, $+14.1\%$, and $+0.4\%$ under severe \emph{label skew} (Dirichlet, Pathological) and \emph{quantity skew}, respectively.
The colon pathology task follows the same trend, with gains of approximately $+13.3\%$, $+9.4\%$, and $+0.2\%$.
Notably, across all datasets under \emph{label skew}, the proposed approach achieves substantial gains of up to $+29.6\%$/$+37.2\%$/$+35.9\%$~(skin lesion/blood cell/colon pathology), which cannot be attributed to trivial update decay.
This consistent pattern across tasks shows that the proposed criterion captures meaningful stabilization under non-IID settings, rather than merely responding to diminishing updates.
Thus, the proposed framework enables reliable hyperparameter tuning across diverse data distributions, with performance comparable to the validation-based stopping.

\subsubsection{Impact of Threshold ($\tau$)}
To investigate the effect of the stopping threshold $\tau$ on the proposed framework, we vary $\tau \in \{0.005, 0.01, 0.05, 0.1\}$ under \emph{label skew} (Dirichlet) with $c=0.1$.
Note that for validation-based early stopping, we report the best test accuracy obtained at the same $\tau$ using either validation loss or validation accuracy.
As shown in Fig.~\ref{fig:result3}, on the skin lesion task, the proposed framework 
shows an overall decreasing trend in test accuracy as $\tau$ increases ($0.005\rightarrow0.1$), with average 
reductions of approximately $19.4\%$ across FL methods.
In particular, FedDyn decreases from $61.0\%$ and $63.9\%$ at $\tau \in \{0.005, 0.01\}$ to $49.6\%$ and $45.8\%$ at $\tau \in \{0.05, 0.1\}$, while FedSMOO decreases from $64.2\%$ to $46.3\%$ ($0.01\rightarrow0.1$).

A similar trend is observed on the blood cell task when compared with validation-based early stopping.
At $\tau \in \{0.005, 0.01\}$, the proposed framework generally matches or exceeds the validation-based baseline across most FL methods, for example achieving $97.2\%$ versus $92.1\%$ for FedSpeed and $86.0\%$ versus $86.1\%$ for FedAvg.
As $\tau$ increases to $0.05$ and $0.1$, the proposed accuracy drops markedly below the validation-based results, reaching $81.1\%$ and $74.6\%$ for FedDyn, and $80.9\%$ and $73.5\%$ for FedSMOO.
On the colon pathology task, the proposed framework similarly surpasses the validation-based baseline at small $\tau$, reaching $92.2\%$ versus $77.7\%$ for FedSAM at $\tau=0.005$, but falls below it at $\tau=0.1$, dropping to $80.2\%$ versus $88.3\%$.
This comparison shows that large $\tau$ values enable fast evaluation but stop training too early to reach the optimum round.
By contrast, small $\tau$ values allow longer federated training, gradually detecting training stabilization and driving the model toward solutions closer to the optimal global model.
Overall, these numerical results establish $\tau$ as a simple and effective control knob for balancing fast evaluation and convergence.

\begin{figure*}[t]
\centering
\includegraphics[width=\textwidth]{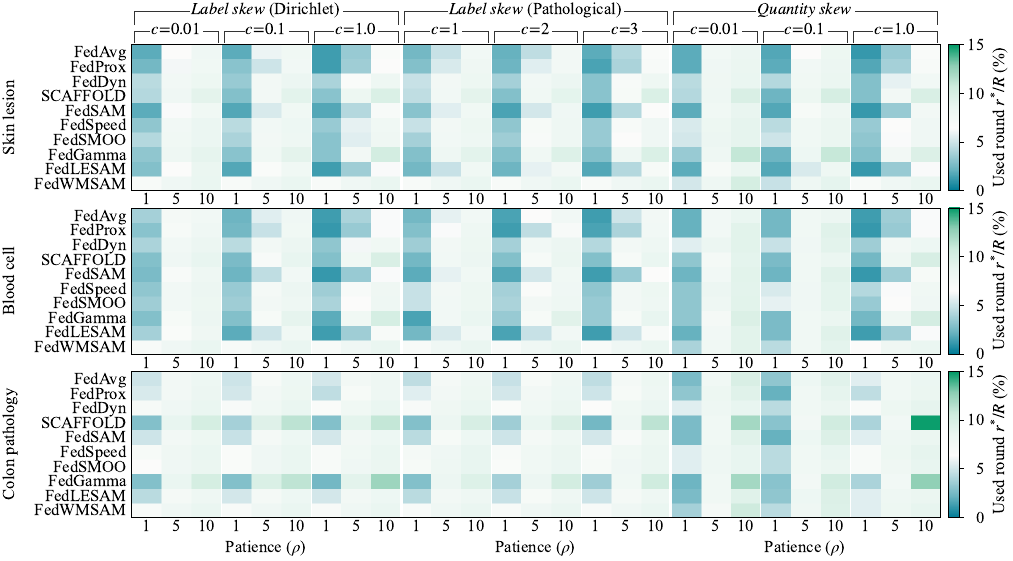}
\caption{Used-round ratio $r^*/R$ (\%) of the proposed framework across FL methods and heterogeneous settings, with $\eta=0.1$ and $\tau=0.1$.
Here, $c \in \{0.01,0.1,1.0\}$ for \emph{label skew} (Dirichlet) and \emph{quantity skew}, $c \in \{1,2,3\}$ for \emph{label skew} (Pathological), and $\rho \in \{1,5,10\}$.
The smaller and larger used-round ratios are denoted by \textbf{\textcolor{myteal}{blue}} and \textbf{\textcolor{mygreen}{green}}, respectively.
}
\label{fig:result4}
\end{figure*}

\subsubsection{Screening of Bad Configurations}
To validate the framework's efficiency, we conduct an ablation study focused on handling bad configurations.
In this scenario, the global model fails to learn and achieves only random-guess level accuracy $14.3\%$/$12.5\%$/$11.1\%$ (skin lesion/blood cell/colon pathology).
As discussed in the threshold analysis, we use a large $\tau$ to enable fast evaluation and early termination.
As shown in Table~\ref{tab:3}, the proposed framework requires only $4$--$34$ additional rounds compared to the best validation-based baseline across all FL methods, far below the fixed budget of $500$ rounds.
On average, the proposed framework stops $\Delta_r \approx +9$ rounds later 
for the skin lesion task and $\Delta_r \approx +8$ rounds later for the blood cell task, 
relative to the best validation-based baseline.
Similarly, the proposed framework stops $\Delta_r \approx +14$ rounds later for the colon pathology task, relative to the best validation-based baseline.
Note that this is less than $3\%$ of the fixed-round budget, enabling rapid screening of bad configurations with minimal overhead.
Therefore, the numerical results demonstrate that large $\tau$ enables efficient resource savings, particularly during early-stage tuning in FL.

\subsubsection{Screening Efficiency under Patience ($\rho$)}
Building on the bad configuration scenario, we vary the patience $\rho$ and measure the used round ratio $r^*/R$ (\%) across diverse non-IID settings, as shown in Fig.~\ref{fig:result4}.
Here, the ratio increases monotonically with $\rho$, as a larger $\rho$ demands more rounds before stopping.
In particular, the average grows from $1.3\%$ at $\rho=1$ to $2.8\%$ and $4.3\%$ at $\rho=5$ and $10$, respectively.
Moreover, this trend is consistent across datasets, where the colon pathology task incurs the highest cost ($1.7\%\rightarrow5.1\%$), while the others remain near $4.0\%$ at $\rho=10$.

Remarkably, even at $\rho=10$, every bad configuration is screened far below the fixed-round budget, peaking at only $14.7\%$ for SCAFFOLD under \emph{quantity skew} ($c=1.0$, colon pathology).
At $\rho=10$ the \emph{quantity skew} setting remains the most demanding at $5.0\%$ on average, slightly above \emph{label skew} (Dirichlet, Pathological) at $4.1\%$ and $3.9\%$.
Indeed, this monotonic dependence persists across all $c$ and \emph{data skew} types, confirming that $\rho$ controls the screening point in the same way as in the validation-based approach.
Thus, these results indicate that $\rho$ serves as a control knob complementary to $\tau$, trading screening speed for robustness at minimal cost.

\section{Conclusion}
\label{sec:conclusion}
In this work, we propose a data-free early stopping framework that identifies the stopping point via global task vector dynamics.
The numerical results show that by tuning the threshold, our framework can either extend training for better performance or match the efficiency of validation-based early stopping.
Moreover, our proposed framework significantly reduces the computational waste of fixed-round FL training by screening ineffective trials.
The proposed approach stops comparably to the validation-based approach, 
requiring only 9/8/14 (skin lesion/blood cell/colon pathology) additional rounds on average for FL methods.
Thus, this work validates the feasibility of data-free early stopping for FL, facilitating practical real-world deployment of FL.

\section*{ACKNOWLEDGEMENTS}
This work was partly supported by the Institute of Information \& Communications Technology Planning \& Evaluation (IITP)-ITRC (Information Technology Research Center) grant funded by the Republic of Korea government (MSIT) (IITP-2026-RS-2020-II201787, contribution rate: 50\%) and (RS-2025-02309685, Development of Programmable Infrastructure Technology for Guaranteed Application Performance, contribution rate: 50\%). In addition, this work was also supported in part by Advanced Micro Devices, Inc. under the AMD University Program’s AI \& HPC Cluster.

\bibliographystyle{IEEEtran}
\bibliography{reference}

\end{document}